\documentclass[twoside,leqno,twocolumn]{article}  
\usepackage{ltexpprt} 

\usepackage{url} 
\usepackage{amsmath}
\usepackage{amssymb}
\usepackage{graphicx}
\usepackage{tikz}
\usetikzlibrary{positioning,patterns}
\usepackage[english]{babel}
\usepackage{booktabs} 
\usepackage[ruled, vlined, linesnumbered, nofillcomment, algo2e]{algorithm2e}
\usepackage{xspace}
\usepackage[tight,center]{subfigure}
\usepackage{color}
\usepackage{caption}
\usepackage{float}
\usepackage{times}

\newcommand{\ourtitle}{Efficiently Summarising Event Sequences\\ with Rich Interleaving Patterns}
\newcommand{\ourmethod}{\textsc{Squish}}

\newcommand{\codeurl}{\url{http://eda.mmci.uni-saarland/squish/}}

\newcommand{\Q}{\ensuremath{Q}}
\newcommand{\Cand}{\ensuremath{W_{\mathit{cand}}\xspace}}
\newcommand{\Sel}{\ensuremath{W_{\mathit{sel}}\xspace}}
\newcommand{\SelX}{\ensuremath{W_{\mathit{add}}\xspace}}
\newcommand{\SelE}{\ensuremath{W_{\mathit{ext}}\xspace}}
\newcommand{\TempX}{\ensuremath{W_{\mathit{tmp}}\xspace}}

\newcommand{\Cp}{\ensuremath{C_p}}
\newcommand{\Cm}{\ensuremath{C_m}}

\newcommand{\code}{\ensuremath{\mathit{code}}}
\newcommand{\CT}{\ensuremath{\mathit{CT}}}
\newcommand{\ST}{\ensuremath{\mathit{ST}}}
\newcommand{\usage}{\ensuremath{\mathit{usage}}}
\newcommand{\fills}{\ensuremath{\mathit{fills}}}
\newcommand{\gaps}{\ensuremath{\mathit{gaps}}}

\newcommand{\length}{\ensuremath{\mathit{length}}}
\newcommand{\gain}{\ensuremath{\mathit{gain}}}
\newcommand{\supp}{\ensuremath{\mathit{supp}}}
\newcommand{\LN}{\ensuremath{L_\mathbb{N}}}

\newcommand{\FINDWIN}{\textsc{Findwin}\xspace}
\newcommand{\GREEDYCOVER}{\textsc{GreedyCover}\xspace}
\newcommand{\ESTIMATE}{\textsc{Estimate}\xspace}
\newcommand{\PRUNE}{\textsc{Prune}\xspace}
\newcommand{\SQS}{\textsc{Sqs}\xspace}
\newcommand{\SQUISH}{\textsc{Squish}\xspace}

\newcommand{\ALIGN}{\textsc{Align}\xspace}
\newcommand{\KRIMP}{\textsc{Krimp}\xspace}
\newcommand{\SLIM}{\textsc{Slim}\xspace}
\newcommand{\GOKRIMP}{\textsc{GoKrimp}\xspace}
\newcommand{\ISM}{\textsc{ism}\xspace}

\newcommand{\SSQS}{\textsc{Sqs-Search}\xspace}

\newcommand{\node}{\mathcal{V}}

\newcommand{\diff}{\mathit{div}}

\makeatletter
\renewcommand*{\@fnsymbol}[1]{\ensuremath{\ifcase#1\or   \circ\or \bullet\or *\or \ddagger\or
   \mathsection\or \mathparagraph\or \|\or **\or \dagger\dagger
   \or \ddagger\ddagger \else\@ctrerr\fi}}
\makeatother

\pgfdeclarelayer{background}
\pgfdeclarelayer{foreground}
\pgfsetlayers{background,main,foreground}

\tikzstyle{block} = [rounded corners, draw=blue!70, fill=white, text width=3.3cm, minimum height=4em]
\tikzstyle{bgblock} = [rounded corners, draw=blue!70, thick, fill=blue!10, text width=3.3cm, minimum height=4em]
\tikzstyle{line} = [draw, -latex', thick,blue!70]

\newcommand{\abs}[1]{{\left|#1\right|}}

\SetKwComment{tcpas}{\{}{\}}
\SetCommentSty{textnormal}
\SetArgSty{textnormal}
\SetKw{False}{false}
\SetKw{True}{true}
\SetKw{Null}{null}
\SetKwInOut{Output}{output}
\SetKwInOut{Input}{input}
\SetKw{AND}{and}
\SetKw{OR}{or}
\SetKw{Continue}{continue}

\makeatletter 
\pgfdeclareshape{event}{ 
\inheritsavedanchors[from=rectangle] 
\inheritanchorborder[from=rectangle] 
\inheritanchor[from=rectangle]{center} 
\inheritanchor[from=rectangle]{north} 
\inheritanchor[from=rectangle]{south} 
\inheritanchor[from=rectangle]{south east} 
\inheritanchor[from=rectangle]{south west} 
\inheritanchor[from=rectangle]{west} 
\inheritanchor[from=rectangle]{east} 
\backgroundpath{
\southwest \pgf@xa=\pgf@x \pgf@ya=\pgf@y 
\northeast \pgf@xb=\pgf@x \pgf@yb=\pgf@y 
\pgf@yc=\pgf@ya \advance\pgf@yc by 2pt 
\pgfpathmoveto{\pgfpoint{\pgf@xa}{\pgf@yc}} 
\pgfpathlineto{\pgfpoint{\pgf@xa}{\pgf@ya}} 
\pgfpathlineto{\pgfpoint{\pgf@xb}{\pgf@ya}} 
\pgfpathlineto{\pgfpoint{\pgf@xb}{\pgf@yc}} 
} 
}
\makeatother

\tikzstyle{tile} = [rounded corners = 2pt, inner sep = 0pt, fill opacity = 0.3, anchor = south west, minimum width = 11pt, minimum height = 7pt]

\newcommand{\gapseq}[2]{%
\foreach \x / \y / \z [count=\xi from 1, count = \xprev from 0]  in {#1}{%
   	\node[tile, right=0 of #2\xprev.south east, anchor = south west, draw=\x, fill=\x!80, minimum width = \y,  text opacity = 1, text = \x, text height=1.4ex,text depth=.5ex] (#2\xi) {\ifthenelse{\equal{\z}{black}}{$\scriptstyle ?$}{$\scriptstyle !$}};%
}%
}

\pgfdeclarelayer{background}
\pgfdeclarelayer{foreground}
\pgfsetlayers{background,main,foreground}

\tikzstyle{block} = [rounded corners, draw=blue!70, fill=white, text width=3.3cm, minimum height=4em]
\tikzstyle{bgblock} = [rounded corners, draw=blue!70, thick, fill=blue!10, text width=3.3cm, minimum height=4em]
\tikzstyle{line} = [draw, -latex', thick,blue!70]

\definecolor{yafaxiscolor}{rgb}{0.3, 0.3, 0.3}

\definecolor{yafcolor1}{rgb}{0.4, 0.165, 0.553}
\definecolor{yafcolor2}{rgb}{0.949, 0.482, 0.216}
\definecolor{yafcolor3}{rgb}{0.47, 0.549, 0.306}
\definecolor{yafcolor4}{rgb}{0.925, 0.165, 0.224}
\definecolor{yafcolor5}{rgb}{0.141, 0.345, 0.643}
\definecolor{yafcolor6}{rgb}{0.965, 0.933, 0.267}
\definecolor{yafcolor7}{rgb}{0.627, 0.118, 0.165}
\definecolor{yafcolor8}{rgb}{0.878, 0.475, 0.686}

\newlength{\yafaxispad}
\newlength{\yaftlpad}
\newlength{\yaflabelpad}
\newlength{\yafaxiswidth}
\newlength{\yafticklen}
\makeatletter
\def\pgfplots@drawtickgridlines@INSTALLCLIP@onorientedsurf#1{}
\makeatother

\begin{document}
\title{\ourtitle}

\author{Apratim Bhattacharyya\thanks{Max-Planck Institute for Informatics and Saarland University, Saarland Informatics Campus, Saarbr\"{u}cken, Germany. \texttt{\{abhattac,jilles\}@mpi-inf.mpg.de}} 
\and Jilles Vreeken\footnotemark[1]}

\date{}

\maketitle

\begin{abstract}
Discovering the key structure of a database is one of the main goals of data mining. In pattern set mining we do so by discovering a small set of patterns that together describe the data well. The richer the class of patterns we consider, and the more powerful our description language, the better we will be able to summarise the data. In this paper we propose \ourmethod, a novel greedy MDL-based method for summarising sequential data using rich patterns that are allowed to interleave. Experiments show \ourmethod is orders of magnitude faster than the state of the art, results in better models, as well as discovers meaningful semantics in the form patterns that identify multiple choices of values.
\end{abstract}

\section{Introduction}\label{sec:intro}

Discovering the key patterns from a database is one of the main goals of data mining. Modern approaches do not to ask for \emph{all} patterns that satisfy a local interestingness constraint, such as frequency~\cite{agrawal:94:fast,mannila:97:discovery}, but instead ask for that \emph{set of patterns} that is optimal for the data at hand. 
There are different ways to define this optimum. The Minimum Description Length (MDL) principle~\cite{rissanen:78:mdl,grunwald:07:book} has proven to be particularly successful~\cite{vreeken:11:krimp,smets:12:slim}. 
Loosely speaking, by MDL we say that the best set of patterns is the set that compresses the data best. How well we can compress, or better, describe the data depends on the description language we use. The richer this language, the more relevant structure we can identify. At the same time, a richer language means a larger search space, and hence requires more efficient search.

In this paper we consider databases of event sequences, and are after that set of sequential patterns that together describe the data best---as we did previously with $\SQS$~\cite{tatti:12:sqs}. Like \SQS we describe a database with occurrences of patterns. Whereas \SQS requires these occurrences to be disjoint, however, we allow patterns to \emph{interleave}. This leads to more succinct descriptions as well as better pattern recall. Moreover, we use a richer class of patterns. That is, we do not only allow for gaps in occurrences, but also allow patterns to emit \emph{one out of multiple} events at a certain location. For example, the pattern `\emph{paper} [\emph{proposes $\mid$ presents}] \emph{new}' discovered in the JMLR abstract database matches two common forms of expressing that a paper presents or proposes something new. 

With this richer language, we can obtain much better compression rates with much fewer patterns. To discover good models we propose \SQUISH, a highly efficient and versatile search algorithm. Its efficiency stems from re-use of information, partitioning the data, and in particular from considering only the currently relevant occurrences of patterns in the data. It is a natural any-time algorithm, and can be ran for any time budget that is opportune.


Extensive experimental evaluation shows that \SQUISH performs very well in practice. It is much better at retrieving interleaving patterns than the very recent proposal by Fowkes and Sutton~\cite{fowkes:16:ism}, and obtains much better compression rates than \SQS~\cite{tatti:12:sqs}, while being orders of magnitude faster than both. 
The choice-patterns it discovers give insight in the data beyond the state of the art, identifying semantically coherent patterns.
Moreover, \SQUISH is highly extendable, allowing for richer pattern classes to be considered in the future.

\section{Preliminaries}\label{sec:prelim}

Here we introduce basic notation, and give short introductions to the MDL principle. 

\subsection{Notation}
We consider databases of \emph{event sequences}. Such a database $D$ is composed of $|D|$ sequences. A sequence $S \in D$ consists of $|S|$ \emph{events} drawn from an alphabet $\Omega$. 
The total number of events occurring in the database, denoted by $||D||$, is simply the sum of lengths of all sequences $\sum_{S \in D} |S|$. We write $S\left[j\right]$ to refer to the $j^{th}$ event in sequence $S$. The support of an event $e \in \Omega$ in a sequence $S$ is simply the number of occurrences of $e$ in $S$, i.e.\ $\supp(e \mid S) = |\{j \mid S[j] = e\}|$. The support of $e$ in a database $D$ is defined as $\supp(e | D) = \sum_{i = 1}^{|D|} supp(e \mid S_{i})$. 

We consider two types of sequential patterns. A \textbf{serial episode} $X \in \Omega^{|X|}$ is
a sequence of $|X|$ events, and we say that a sequence $S$ contains $X$ if
there is a subsequence in $S$ equal to $X$. We allow noise, or
\emph{gap events}, within an occurrence of $X$.
We also consider choice episodes, or \textbf{choicisodes}. These are serial episodes with positions matching \emph{one out of multiple} events. For example, serial episode $ac$ matches an $a$ followed by $c$, whereas choicisode $\left[ a , b \right] c$ matches occurrences of $a$ or $b$ followed by $c$.

\subsection{Brief introduction to MDL}\label{mdl}
The Minimum Description Length principle (MDL) \cite{rissanen:78:mdl,grunwald:07:book} is a practical version of Kolmogorov Complexity~\cite{vitanyi:93:book}. Both embrace the slogan Induction by Compression. We use the MDL principle for model selection.
 
By MDL, the best model is the model that gives the best lossless compression. More specifically, given a set of models $\mathcal{M}$, the best model $M \in \mathcal{M}$ is the one that minimizes $L(M)$ + $L(D \mid M)$, in which $L(M)$ is the length in bits of the description of $M$, and $L(D \mid M)$ is the length of the data when encoded with model $M$. Simply put, we are interested in that model that best compresses the data without loss.
MDL as describe above is known as two-part MDL, or crude MDL; as opposed to refined MDL. In refined MDL model and data are encoded together \cite{grunwald:07:book}. We use two-part MDL because we are specifically interested in the model: the patterns that give the best description. 
In MDL we are only concerned with code lengths, not actual code words.

Next, we formalise our problem in terms of MDL.

\section{MDL for Event Sequences}\label{sec:score}

To use MDL we need to define a model class $\mathcal{M}$, and how to encode a model $M \in \mathcal{M}$ and data $D$ in bits.

As models we will consider \emph{code tables}~\cite{vreeken:11:krimp,tatti:12:sqs}. A code table $\CT$ is a dictionary between patterns and associated codes. A code table consists of the singleton patterns $e \in \Omega$, as well as a set $\mathcal{P}$ of non-singleton patterns. We write $\code_{p}(X)$ to denote the \emph{pattern code} that identifies a pattern $X \in \CT$. 
Similarly, we write $\code_f(X)$ and $\code_g(X)$ for the codes resp.\ identifying 
a \emph{fill} resp.\ a \emph{gap} in the occurrence of a pattern $X$. 

\begin{figure}[bt!]
\begin{center}
\newlength{\tilelen}
\addtolength{\tilelen}{11pt}

\includegraphics{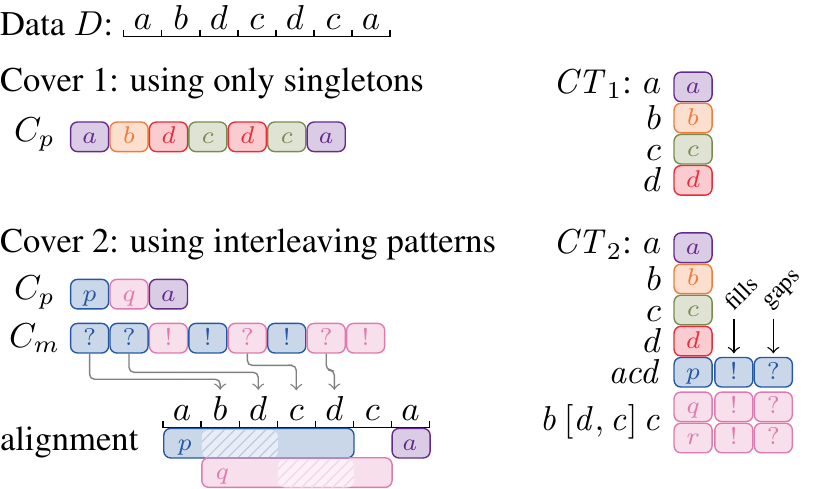}
\end{center}
\caption{Example of two possible covers. The first uses only singletons, the second includes interleaving and choicisodes.}
\label{fig:covers}
\end{figure}

We can encode a sequence database $D$ using the patterns in a code table $\CT$, which generates a \emph{cover} $C$ of the database. A cover $C$ uniquely defines a pattern code stream $C_p$ and a meta code stream $C_m$. The pattern stream is simply the concatenation of the codes corresponding to the patterns in the cover, in the order of their appearance. Likewise, the meta stream $C_m$ is the concatenation of the gap and fill codes corresponding to the cover. In Fig.~\ref{fig:covers}, we illustrate two example covers and corresponding code tables, the first using only singletons and the second cover with interleaving using patterns from a richer code table with choicisodes.

Before formalising our score, it is helpful to know how to decode a database given a code table and the code streams. 

\subsection{Decoding a database}

To decode a database, we start by reading a $\code_p(X)$ from the pattern stream $\Cp$. If the corresponding pattern $X$ is a singleton, we append it to our reconstruction of the database $D$. If it is a non-singleton, we append its first event, $X[1]$, $D$. To allow for interleaving, we have to add a new \emph{context} to context list $\Lambda$. A context is a tuple $(X,i)$ consisting of a pattern $X$, and a pointer $i$ to the next event to be read from the pattern. For an example, let us consider Cover 2 in Fig.~\ref{fig:covers}. We read $\code_p(p)$ from $\code_p$, append $p[1]=a$ to $D$, and add $(p,2)$ to the context list.

Next, if the context list is non-empty, we read as many meta codes from $C_m$ as there are contexts in $\Lambda$. 
If we read a fill code $\code_f(X)$ corresponding to one of the contexts $(X,i) \in \Lambda$, we append the next event from $X$, $X[i]$ to the data $D$, and increment the pointer. If after this step we have finished reading the pattern, we remove its context from the list. If we only read gap codes $\code_g(X)$ for every pattern $X$ in the context list, we read again from the pattern stream. We do this until we reach the end of the pattern stream $\Cp$.

Continuing our example, we read $\code_g(p)$ from $C_m$, which corresponds to a gap in the occurrence of pattern $p$. We read $\code_p(q)$ from $C_p$, write $q[1]=b$ to $D$, and insert context $(q,2)$ to $\Lambda$. Next, $\Lambda$ contains two contexts, and we read two meta codes from $C_m$, viz.\ $\code_g(p)$ and $\code_f(q)$. As for context $(q,2)$ we read a fill code, we write $q[2]=d$ to $D$, and increment its pointer to $3$. Etc.

\subsection{Calculating Encoded Lengths}\label{comenlen}
Given the above scheme we know which codes to expect when, and can now formalise our score. We build upon and extend the encoding on Tatti \& Vreeken~\cite{tatti:12:sqs} for richer covers and patterns.

\subsubsection*{Encoded Length of the Database}
We encode the pattern stream $\Cp$ using Shannon optimal prefix codes. The length of the pattern code  $L(\code_{p}(X))$ for a pattern $X$ depends on how often it is used in the pattern stream. We write $\usage(X)$ to denote the number of times $\code_{p}(X)$ occurs in $\Cp$. The length of the optimal pattern code for $X$ then is
\begin{align*}
L(\code_{p}(X)) &= - \log\Big(\frac{\usage(X)}{\sum_{Y \in \CT} \usage(Y)}\Big)\,.
\end{align*}
The encoded length of the whole pattern stream $C_p$ is then simply $L(C_p) = \sum_{X \in \CT} \usage(X) L(\code_p(X))$.

To avoid arbitrary choices in the model encoding, we use prequential codes~\cite{grunwald:07:book} to encode the meta stream. Prequential codes are asymptotically optimal without knowing the distribution beforehand. The idea is that we start with an uniform distribution over the events in the stream and update the counts after every received event. This means we have a valid probability distribution at every point in time, and can hence send optimal prefix codes. The total encoded length of the meta stream pattern is
\begin{align*}
L(\Cm) =& \sum_{X \in CT} \Big( -\sum_{i=1}^{\fills(X)}\log\left(\frac{\epsilon + i}{ 2 \epsilon + i }\right) \\
& -\sum_{i=1}^{\gaps(X)}\log\left(\frac{\epsilon + i}{ 2 \epsilon  + \fills(X) + i }\right) \Big) \,.
\end{align*} 
where $\epsilon=0.5$ is a constant by which we initialize the distribution~\cite{grunwald:07:book}, $\fills(X)$ and $\gaps(X)$ are the number of times $\code_f(X)$ resp.\ $\code_g(X)$ occurs in $C_m$. 

For lossless decoding of database $D$, the number of sequences $|D|$ and the length of each sequence $S \in D$ should also be encoded. We do this using $\LN$, the MDL optimal code for integers $n \geq 1$~\cite{rissanen:83:integers}. 

Combining the above, for the total encoded length of a database, given a code table $\CT$ and cover $C$, we have
\[
L(D \mid \CT) = \LN(| D |) + \sum_{S \in D} \LN(| S |)  + L(\Cp) + L(\Cm)\, .
\]
Next we discuss how to encode a model.

\subsubsection*{Encoded Length of the Code Table}\label{comenlen}
Note that the simplest valid code table consists of only the singletons $\Omega$. We refer to this code table as $\ST$, or, the standard code table. We use $\ST$ to encode the non-singleton patterns $\mathcal{P}$ of a code table $\CT$. 
The usage of a singleton $e \in \ST$ is simply its support in $D$, and hence the code length $\code_p(e) =  - \log\Big( \frac{\supp(e \mid D)}{||D||} \Big)$.
To use these codes the recipient needs to know the supports of the singletons. We encode these using a data to model code---an index over a canonically ordered enumeration of all possibilities~\cite{vereshchagin:03:kolmo}; here it is the number of possible supports of $| \Omega |$ alphabets over a database length of $|| D ||$, $\binom{|| D ||}{|\Omega |}$. The length of the code is now simply the logarithm over the number of possibilities.

Given the standard code table $\ST$, we can now encode the patterns in the code table. We first encode the length $|X|$ of the  pattern, and then number of choice spots in the pattern, $||X|| - |X|$. We encode how many choices we have per location using a data to model code. We finally encode the events $X[i]$ using the standard code table, $\ST$. That is, 
\begin{align*}
L(X \mid \ST) &= \LN(|X|) + \LN( || X || - |X| + 1)\\
&+ \log \genfrac{(}{)}{0pt}{}{||X||-1}{|X| - 1} + \sum_{i = 1}^{|| X ||} L(X[i] \mid \ST) \, .
\end{align*}
Note that if we do not consider choicisodes, we can simplify the above as we only need to transmit the first and last part of this code. That is, the length and the events in the pattern.

Recall that, pattern codes in the pattern stream $\Cp$ are optimal prefix codes. The occurrences of the non-singleton patterns need to be transmitted with the model. We do this again using a data to model code. We encode the sum of pattern usages, $\usage(\mathcal{P}) = \sum_{X \in \CT \setminus \Omega } \usage(X)$, by the MDL optimal code for integers. It is equivalent to use a pattern code per choicisode and then identify the choice-events, or to use a separate pattern code for each instantiation of the choicisode. For simplicity we make the latter choice. 

The total encoded size of code table $\CT$ given a cover $C$ of database $D$ is then given by
\begin{align*}
L(\CT \mid D, C) =& \LN(| \Omega |) + \log {||D|| \choose |\Omega|} \\
&+ \LN(| \mathcal{P} | + 1) + \LN(\usage(\mathcal{P}) + 1) \\
&+ L(\usage(\mathcal{P}),| \mathcal{P} |) + \sum_{X \in CT}L(X,\CT)\, .
\end{align*}

We are interested in the set of patterns and a corresponding cover $C$ which minimizes the total encoded length of the code table and the database, which is,
\begin{align*}
L(\CT,D) &= L(\CT \mid C) + L(D \mid \CT)\, .
\end{align*}
We can now formally define our problem as follows. 

\vspace{0.5em}\noindent{\textbf{Minimal Code Table Problem}} Let $\Omega$ be a set of events and let $D$ be a sequence database over $\Omega$, find the minimal set of serial (choice) episodes $\mathcal{P}$ such that for the optimal cover $C$ of $D$ using $\mathcal{P}$ and $\Omega$, the total encoded cost $L(\CT , D)$ is minimal, where $\CT$ is the code-optimal code table for $C$.
\vspace{0.5em}

For a given database $D$, we would like to find its optimal pattern set in polynomial time. However, there are exponentially many possible pattern sets, and given a pattern set, there are exponentially many possible covers. 
For neither problem there exists trivial structure such as monotonicity or sub-modularity that would allow for an optimal polynomial time solution. 

Hence, we resort to heuristics. In particular, we split our problem into two parts. We first explain our greedy algorithm to find a good cover given a set of patterns. We describe how to find a set of good patterns in Sec.~\ref{sec:search}.

\section{Covering a Database}\label{sec:cover}
Given a pattern set $\mathcal{P}$ and database $D$, we are after a cover $C$ with interleaving and nesting, that minimises $L(\CT, D)$. 

Each occurrence of a pattern $X$ in database $D$, possibly with gaps, defines a \emph{window}. We denote by $S\left[ a,b \right]$ a window in sequence $S$ that extends from the position $a$ to $b$. Two windows are non-overlapping if they do not have any events in common which belong to their respective patterns. Two interleaving or nesting windows might have common events, which, as we do not allow overlap, leads to gap events for one of the two windows. Two windows are disjoint if they do not have any events in common. For every event in the database $D$, there can be many windows with which we can choose to cover it. The optimal cover depends upon the pattern, fill, gap codes of the patterns. The choices grow exponentially with sequence length, with no trivial sub-structure. 

To find good disjoint covers, Tatti \& Vreeken~\cite{tatti:12:sqs} use an EM-style approach. At each step until convergence, given the pattern, gap and fill codes, the authors use the dynamic programming based algorithm $\ALIGN$ to find a cover. $\ALIGN$ takes a set of possibly overlapping minimal windows and returns a subset of disjoint minimal windows (i.e.\ a cover) which maximizes the sum of $\gain$ (a heuristic measure) of each window. Then, the lengths of the codes are reset based upon the found cover. It is unclear if this scheme can be extended to return a cover with interleaved or nested windows efficiently. Moreover if we extend our model with a new pattern, we have to rerun $\ALIGN$ from scratch. 

We propose an efficient and easily extendible heuristic for good covers with interleaved and nested windows. 

\subsection{Window Lengths}
For a given pattern, as we consider windows with gaps, the length of an window in the database can be arbitrarily long. Tatti \& Vreeken therefore consider only \emph{minimal windows}. A window $w = S\left[ i, j \right]$ is a minimal window of a pattern $X$ if $w$ contains $X$ but no other proper sub-windows of $w$ contain $X$. If no interleaving or nesting is allowed, it is optimal to consider only minimal windows. Otherwise, it is easy to construct examples where the optimal cover consists of non-minimal windows.

\begin{figure}[bt!]
\begin{center}
\includegraphics{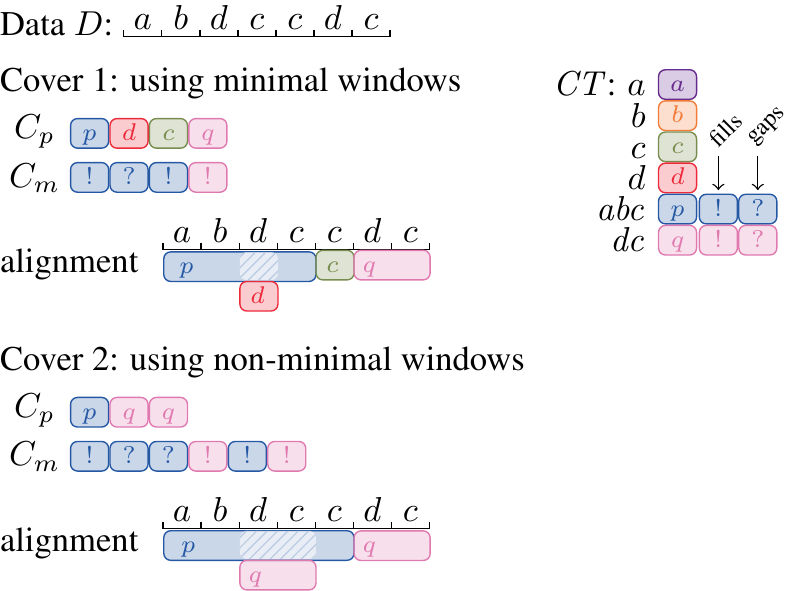}
\end{center}
\caption{Two possible covers using the same code table. The first cover uses only minimal windows. The second includes nesting and a non-minimal window of \emph{abc}. The second cover leads to an overall lower encoded length. }
\label{fig:nonmin}
\end{figure}

Consider the sequence \emph{abdccdc} and a code table with the pattern \emph{abc}, \emph{dc} and the singletons \emph{a}, \emph{b}, \emph{c} and \emph{d}. Two possible covers are: \emph{(\underline{ab} d \underline{c}) c \underline{dc}} using only minimal windows and \emph{(\underline{ab(dc)c}) \underline{dc}} where a non-minimal window of \emph{abc} is used and is nested with a window of \emph{dc}. It is easy to see that the second cover leads to lower encoded length $L(abdccdc, \left\{ abc, dc, a, b, c, d\right\} )$ (see Fig \ref{fig:nonmin}) of about 2.9 bits.

Ideally, we should consider all possible windows. The number of possible windows of a pattern, however, is quadratic in the length of the database. This means that even a search for all windows is computationally inefficient. Therefore, we first search for only the shortest window from each starting position in the database. We consider longer windows when necessary. We do so as follows.

\subsection{Window Search}

Given a pattern $X$, we use the pseudo-code $\FINDWIN$ presented as Algorithm \ref{findwin} to search for its windows in the a sequence or sub-sequence $S$ of database $D$. It returns us $\Cand(X)$ which is a set of candidate windows of the pattern $X$. It considers only the first window from each starting position in the sequence $S$. We later choose a subset of these windows (along with those of patterns other than $X$ in $\CT$) to create a cover of the database $L(D \mid \CT)$. To control the ratio of gaps and fills, we maintain a budget variable. This is the number of extra allowed overall gaps. Ideally we would like to have more fills than gaps as it leads to better compression.

To search for windows efficiently in $\FINDWIN$, we use an inverted index: $\mathit{index}^{-1}(x)$ which gives us a list of positions of the event $x \in \Omega$ in the database. We use a priority queue $\Q$ to store potential windows sorted by length. Shorter windows means more fills than gaps. We initialize $\FINDWIN$ (line 3) by creating potential windows at all the positions where the first alphabet of pattern $X$ occurs in the sequence $S_{i}$ and pushing these potential windows to $\Q$. Each window $w$ in $\Q$ contains the starting position in $S_{i}$, its length, and a pointer $w_{i}$. This pointer points to a certain event in pattern $X$ which we are search for in $S_{i}$. At every step of $\FINDWIN$ (line 7) we look at the potential window at the top of the queue $\Q$. We check if the next event in the database equals the character of the pattern $X$ pointed to by $w_{i}$ and increment the length of the window $w$. There are now two possibilities i) (line 10) The next database event is the same as the event in $X$ pointed to by $w_{i}$.  If we have found the full pattern $X$ in the database, we add this window to $\Cand(X)$. We can now update our budget if we used more fills than gaps. Using less gaps in one window allows us to use more gaps in another. ii) (line 15) The next database event does not equal the event of $X$ pointed to by $w_{e}$. This means that the potential window $w$ has one extra gap. We check if this extra gap is allowed by our budget (line 16). Otherwise, we drop the window.

\begin{algorithm2e}[t]
\small
 \Input{sequence $S$ and a pattern $X$}
 \Output{set of windows for $X$}
 $\Cand \leftarrow \emptyset$\;
 \For{$p$ in $\mathit{index}^{-1}(x)$}{
 $w \leftarrow (\mathit{start}, \mathit{length}, w_{i} = 1)$\;
 $\mathit{push}(w,\Q)$\;
 }
 \While{$T$ is not empty}{
 $w \leftarrow \mathit{top}(\Q)$\;
 $(\mathit{start}, \mathit{length}, w_{i}) \leftarrow w$\;
 $\mathit{length} = \mathit{length} + 1$\;
 \eIf{$S_{i}\left[ start + length \right]$ equals $X\left[w_{i}\right]$}{
	\If{$w_{i}$ points to the end of $X$}{
	append $w$ to $\Cand$\;
	$b \leftarrow b + 2 \times \length(X) - \mathit{length} - 1$\;
	}
 }
{
 \If{$b + 2 \times \length(X) - \mathit{length} - 1 < 0$}{
	delete $w$ from $\Q$\;
	}
 }
 }
 \caption{$\FINDWIN(S_{i},X,budget)$}\label{findwin}
\end{algorithm2e}

Now that we can search for windows of patterns, we describe how to choose a subset which generates a good cover $C$ of the database $D$.

\subsection{Candidate Order}  
In the first step of our greedy strategy, we sort the set of patterns in a fixed order, similar to \cite{bertens:16:ditto}. We call this order the \textbf{Candidate Order}. We cover the database using windows of patterns in this order. This order is designed to minimize the code length. This is achieved by putting longer and more frequently occurring patterns higher up in the candidate order. This means we can cover more events while minimizing the code length.

We consider the patterns $X \in \CT$ in the order,
\begin{enumerate}
\item Decreasing $\downarrow$ in length $\mid X \mid$
\item Decreasing $\downarrow$ in support $support(X \mid D)$
\item Decreasing $\downarrow$ in length of encoding it with the standard table.
\item Increasing $\uparrow$ lexicographically.
\end{enumerate}

\subsection{Greedy Cover}\label{greedyc}
We now describe our greedy algorithm $\GREEDYCOVER$ which we use incrementally build a good cover as pseudo-code in Algorithm \ref{coverint}. We consider patterns in the candidate order. We maintain a set of selected windows $\Sel$. $\GREEDYCOVER$ takes this set of selected windows $\Sel$ and extends it with a subset of candidate windows of pattern $X$, $\Cand(X)$, found with $\FINDWIN$ and possibly with (longer, interleaved) windows found on the fly. We assume that both $\Sel$ and $\Cand$ are sorted.

We refer to a block of windows which are interleaved or nested with each other as an \emph{window extend}. For ease of notation, we refer to windows which are not interleaved or nested also as window extends (containing a single window). We begin $\GREEDYCOVER$ by dividing the set $\Sel$ into a set of window extends $\SelE$ by a linear sweep (if $\Sel$ is sorted). For patterns at the top of the candidate order $\Sel$ is empty, so we can select all candidate windows $\Cand(X)$. For any other pattern, we iterate though the list of window extends $\SelE$ (line 4). All the windows of the pattern occurring between any two extends in $\SelE$ can be potentially chosen. These are put in $\TempX$ (line 5), a temporary list. It is possible that some windows of $X$ in $\TempX$ overlap. We consider these windows in order of decreasing length (line 6) and discard any window that overlaps with a previously chosen window. We additionally search (on the fly) for interleaved windows occurring within the window extends $\SelE$ (line 9). 

For example consider the sequence \emph{abcdacbd} which we want to cover with the patterns \emph{ac} and \emph{bd}. Using $\FINDWIN$ we get two windows each for the two patterns. If \emph{ac} is higher up in the candidate order, we first select the two windows of \emph{ac}; \emph{\textbf{a}b\textbf{c}d\textbf{ac}bd}. We now have two window extends in $\SelE$. We search for windows of \emph{bd} within the first window extend of \emph{ac} to find one interleaved window: \emph{a\textbf{b}c\textbf{d}acbd} and we select the second window of \emph{bd} as it is between the  two window extends of \emph{ac}.

\begin{algorithm2e}[t]
\small
 \Input{set of selected windows $\Sel$ and a set of candidate windows $\Cand(X)$ for pattern $X$.}
 \Output{set of selected windows $\Sel$ combined with those in $\Cand(X)$ not overlapping with $\Sel$.}
 $\SelX \leftarrow \emptyset$\;
 $\SelE \leftarrow $ create window extends from $\Sel$ and $\Cand(X)$\;
 $\mathit{last} \leftarrow \emptyset$\;
 \For{window $w \in \SelE$}{
 $\TempX \leftarrow$ all windows of $X$ between $\mathit{last}$ and $v$\; 
 \For{$v$ in $\TempX$, in order of decreasing length}{
 \If{$v$ does not overlap with $\TempX$}{
 append $v$ to $\SelX$\;
 }
 }
 $\TempX \leftarrow \TempX \cup \{$ windows of $X$ inside $\mathit{w}$ in $D\}$\;
 $\mathit{last} \leftarrow \mathit{w}$\;
 }
 \Return Merge $\Sel$ and $\SelX$\;
 \caption{$\GREEDYCOVER(\Sel,\Cand)$}\label{coverint}
\end{algorithm2e} 

Note that, $\GREEDYCOVER$ now takes time $O(|\Sel| + ||D|| + |\Cand| \log(|\Cand|))$, in the worst case. Where, $|\Sel|$ is the number of windows in $\Sel$ and $|\Cand|$ is the number of candidate windows. Let, $W_{\max(\mathcal{P})}$ be the maximum number of candidate windows of any pattern in $\mathcal{P}$. Then $\GREEDYCOVER$ takes time $O( |\mathcal{P}| \, ( W_{\max(\mathcal{P})} \log(W_{\max(\mathcal{P})}) + ||D|| ))$ to construct a cover $C$ of the database $D$ using the patterns $\mathcal{P}$ in the code table $\CT$ in the worst case. The maximum number of candidate windows of any pattern $W_{\max(\mathcal{P})}$ is bounded by the size of the database $O(||D||)$. However, $\GREEDYCOVER$ makes it computationally more efficient to extend the code table with a new pattern $X$. We can discard windows of patterns in $\mathcal{P}$ below $X$ in the candidate order from the cover and run $\GREEDYCOVER$ for $\Cand(X)$ and the patterns in the code table below $X$ in candidate order. This means that we do not have to recompute the cover from scratch. This is very efficient if the pattern $X$ is near the bottom of the candidate order. As we shall see $\GREEDYCOVER$ is very competitive in its execution time compared to \SQS~\cite{tatti:12:sqs}.

Having presented our greedy approach of covering a database given a set of patterns, we now turn our attention to the task of mining good set patterns.

\section{Mining Good Code Tables}\label{sec:search}
Given a pattern set $\mathcal{P}$ we have a greedy algorithm to cover the database $D$ and obtain the encoded length of the model and data $L(\CT,D)$. To solve the \textbf{Minimal Code Table Problem} we want to find that set $\mathcal{P}$ of patterns which minimizes the total encoded length $L(\CT,D)$ of the database. As discussed before, there does not seem to be any trivial sub-structure in the problem which we can exploit to obtain an optimal set of patterns $\mathcal{P}$ in polynomial time. So, we resort to heuristics. We build upon and extend $\SSQS$~\cite{tatti:12:sqs}.

\subsection{Generating Candidates}\label{subsec:gencan}
We build a pattern set $\mathcal{P}$ incrementally. Given a set of patterns $\mathcal{P}$ and a cover $C$, we aim to find a pattern $X$ and an extension $Y$, such that $X,Y \in \mathcal{P} \cup \Omega$, whose combination $XY$ would decrease the encoded length $L(\mathcal{P} \cup XY,D)$. We do this until we cannot find any $XY$ that when added to $\mathcal{P}$ reduces the total encode size. Doing is exactly, however, is computationally prohibitive. At every iteration, there would be $O((|\mathcal{P}| + |\Omega|)^{2})$ possible candidates. Thus, we again resort to heuristics. We use the heuristic algorithm $\ESTIMATE$ from \cite{tatti:12:sqs} that can find good candidates, with likely decrease in code length if added, in $O(| \mathcal{P} | + | \Omega| + ||D||)$  time. 
For readability and succinctness, we 
describe algorithm $\ESTIMATE$ in
Appendix~\ref{sec:apx}.

Candidates are accepted or rejected based on the compression gain. As we can now find richer covers with interleaving and nesting, candidates are potentially more likely to be accepted. However, we want to find a succinct set of patterns which describe the data well. Choicisodes can help in this search for a succinct summary of the data.  

\subsection{Choicisodes}
Recall from Sec.~\ref{comenlen} that we can encode patterns as choicisodes. We have the possibility of combining a newly discovered non-singleton pattern with a previously discovered non-singleton pattern or choicisode to create or expand a choicisode. Combining non-singleton patterns into a single choisisode may hence lead to savings in the encoded length of the code table $L(\CT \mid C)$ while providing a more succinct representation of the pattern set.

We use a greedy strategy based on MDL for discovering choicisodes. For each newly discovered non-singleton pattern, we consider all previously discovered non-singleton patterns or choicisodes which differ with it at one position. Then, we calculate the increase in code length (of the model) if we encode it as a choicisode with each of these non-singleton patterns or choicisodes. We also consider the increase in code length if we encode it as independently. We choose whichever option with leads to the minimum increase in code length.

Next we present our algorithm $\SQUISH$ for mining a succinct and representative pattern set.

\subsection{The SQUISH algorithm}
The present the complete algorithm $\SQUISH$ as pseudo-code in Algorithm \ref{search}. At each iteration, it considers each pattern $X \in \CT$. It creates potential extensions $XY$, with $Y \in \CT$, based on estimated change in the encoded length using $\ESTIMATE$ (line 6). $\SQUISH$ then considers each of these patterns in the order of the estimated decrease in gain if added to $\CT$ (line 7). $\FINDWIN$ is used to find the candidate windows of each of these extensions (line 9). $\GREEDYCOVER$ is used to cover the data with this candidate pattern $XY$ added to $\CT$. We simultaneously consider the possibility of encoding $XY$ as a choicisode. If $XY$ leads to a decrease in the encoded length of the database $D$ then, we add $XY$ to $\CT$. If $XY$ is to be added, we $\PRUNE$ 
(see Appendix~\ref{sec:apx}) the code table to remove redundant patterns. Consider, for example if we decide to add \emph{abcd}, the pattern \emph{ab} and \emph{cd} may not be required to construct an effective cover of the database. We also consider the singletons occurring in the gaps of $XY$, by constructing new extended patterns by using these gap alphabets as intermediate alphabets.

\begin{algorithm2e}[t]
\small
 \Input{database $D$}
 \Output{pattern set $\mathcal{P}$ with low $L(\CT,D)$}
 $\mathcal{P} \leftarrow \phi$\;
 $C \leftarrow \GREEDYCOVER(\mathcal{P},D)$\;
 \While{\emph{changes}}{
 $\mathcal{F} \leftarrow \phi$\;
 \For{$X \in \CT$}{ add $\ESTIMATE(X, A, D)$ to $\mathcal{F}$\;} }
 \For{ $Z \in \mathcal{F}$ ordered by estimated gain}{
    Sort $\mathcal{P} \cup Z$ in candidate order\;
    $\Cand(Z) \leftarrow \FINDWIN(D, Z, \mathit{budget})$\;
    $C \leftarrow \GREEDYCOVER(\mathcal{P} \cup Z,D)$\;
 	\If{ $L(D, \mathcal{P} \cup Z) < L(D, \mathcal{P})$ } { 
 	$\mathcal{P} \leftarrow \PRUNE(\mathcal{P} \cup Z, D)$\;
 	}
 }
 \caption{$\SQUISH(D)$}\label{search}
\end{algorithm2e}  

\section{Related Work}\label{sec:related}
Discovering sequential patterns is an active research topic. Traditionally there was a focus on mining frequent sequential patterns, with different definitions of how to count occurrences~\cite{mannila:97:discovery,wang:04:bide,laxman2007fast}. Mining general patterns, patterns where the order of events are specified by a DAG is surprisingly hard. Even testing whether a sequence contains a pattern is NP-complete~\cite{tatti:11:multievent}. Consequently, research has focused on mining subclasses of episodes, such as, episodes with unique labels \cite{achar2012discovering,pei2006discovering}, strict episodes \cite{tatti:12:clsepi}, and injective episodes~\cite{achar2012discovering}.

Traditional pattern mining typically results in overly many and highly redundant results. Once approach to counter this is mining statistically significant patterns. Computing the expected frequency of a sequential pattern under a null hypothesis is very complex, however~\cite{tatti:15:epirank,petitjean:16:skopus}. 

\SQUISH builds upon and extends \SQS~\cite{tatti:12:sqs}. Both draw inspiration from the \KRIMP~\cite{vreeken:11:krimp} and \SLIM~\cite{smets:12:slim} algorithms. \KRIMP pioneered the use of MDL for mining good patterns from transaction databases. Encoding sequential data with serial episodes is much more complicated, and hence \SQS uses a much more elaborate encoding scheme. Here, we extend it to discover richer structure in the data. 
The \SLIM algorithm~\cite{smets:12:slim} mines \KRIMP code table directly from data. \SLIM iteratively seeks to improve the current model by considering as candidates joins $XY$ of patterns $X,Y \in \CT$. Whereas \SLIM considers the full Cartesian product and ranks on the basis of estimated gain, \SQS and \SQUISH take a batch based approach.

Lam et al.\ introduced \GOKRIMP \cite{lam:12:gokrimp} for mining sets of serial episodes. As opposed to the MDL principle, they use fixed length codes, and do not punish gaps within patterns. This means, their goal is essentially to cover the sequence with as few patterns as possible, which is different from our goal of finding patterns that succinctly summarize the data. 

Recently, Fowkes and Sutton proposed the $\ISM$ algorithm \cite{fowkes:16:ism}. \ISM is based on a generative probabilistic model of the sequence database, and uses EM to search for that set of patterns that is most likely to generate the database. \ISM does not explicitly consider model complexity. Like \SQUISH, \ISM can handle interleaving and nesting of sequences. We will empirically compare to \ISM in the experiments.

\section{Experiments}
Next we empirically evaluate \SQUISH on synthetic and real world data. We compare against \SQS~\cite{tatti:12:sqs} and \ISM~\cite{fowkes:16:ism}. 
All algorithms were implemented in C++. We provide the code for research purposes.\!\footnote{\codeurl} 

We evaluate quantitatively on the basis of achieved compression, pattern recall, and execution times. Specifically, we consider the compression gain $\Delta L = L(D,\ST) - L(D,\CT)$. That is, the gain in compression using discovered patterns versus using the singleton-only code table. Higher scores are better. 
All experiments were executed single threaded on quad-core Intel Xeon machines with 32GB of memory, running Linux.

\subsection*{Databases}
We consider four synthetic, and five real databases. We give their base statistics in Table \ref{dbstat}. 

\begin{table}
\centering
\caption{Database Statistics}
\label{dbstat}
\begin{tabular}{l rrrr}
\toprule
\textbf{Dataset} & $|\Omega|$ & $|D|$ & $||D||$ & $L(D,\ST)$  \\
\midrule
Indep & 1k & 1 & 10k & 103\,630\\
Plant-10 & 1k & 1 & 10k & 103\,340\\
Plant-50 & 1k & 1 & 10k & 102\,630\\ 
Parallel & 25 & 10k & 1M & 4\,644\,290 \\
\midrule
Sign & 267 & 730 & 38\,689 & 271\,232 \\
Gazelle & 497 & 59k & 209\,240 & 1\,179\,030 \\
Address & 5\,295 & 56 & 62\,066 & 685\,593 \\
JMLR & 3\,846 & 788 & 75\,646 & 772\,112 \\
Moby & 10\,277 & 1 & 105\,719 & 1\,250\,149 \\
\bottomrule
\end{tabular}
\end{table}

\textit{Indep}, \textit{Plant-10}, and \textit{Plant-50} are synthetic data consisting of a single sequence of 10\,000 events, over an alphabet of 1000 events. For \textit{Indep}, all events are independent. For \textit{Plant-10}, and \textit{Plant-50} we plant resp.\ 10 and 50 patterns of 5 events long 10 times each over an otherwise independent sequence, with a 10\% probability of having a gap between consecutive events. To evaluate the ability of \SQUISH to discover interleaved and nested patterns, we consider the \textit{Parallel} database~\cite{fowkes:16:ism}. Each event in this database is generated by five independent parallel processes chosen at random. Each process $i$ generates the events $\left\{ a_{i}, b_{i}, c_{i}, d_{i}, e_{i} \right\}$ in sequence. 

\begin{table*}
\footnotesize
\centering
\caption{Comparing \SQS and \SQUISH. Results for \SQUISH using resp. disjoint serial episodes, interleaving serial episodes, and interleaving serial episodes and choicisodes. Given are the number of non-singleton patterns ($|\mathcal{P}|$), time to finish ($t$), time reach the same score as \SQS ($\SQS$-$t$), and the gain in compression $\Delta L$ (higher is better).\label{tbl:dcresults}}
\begin{tabular}{l rrr rrrr rrrr rrrr}
\toprule
&&&& \multicolumn{8}{l}{$\SQUISH$} \\
\cmidrule{5-16}
& \multicolumn{3}{l}{$\SQS$} &
\multicolumn{4}{l}{Disjoint} & 
\multicolumn{4}{l}{Interleaving} & 
\multicolumn{4}{l}{Choisisodes}\\
\cmidrule(r){2-4}
\cmidrule(lr){5-8}
\cmidrule(lr){9-12}
\cmidrule(l){13-16}
\textbf{Dataset} & $|\mathcal{P}|$ & $t$ & $\Delta L$ & $|\mathcal{P}|$ & $\SQS$-$t$ & $t$ & $\Delta L$ & $|\mathcal{P}|$ & $\SQS$-$t$ & $t$ & $\Delta L$ & $|\mathcal{P}|$ & $\SQS$-$t$ & $t$ & $\Delta L$ \\
\midrule
Sign & 127 & 81s & 15.5k & 
	157 &  \textbf{3.0s} & 59s & \textbf{22.5k} & 
	156 & \textbf{4.3s} & 132s & \textbf{22.7k}& 
	93 & \textbf{4.3s} & 103s & \textbf{23.5k} \\
Gazelle & 934 & 26m & 14.7k & 
	880 &  \textbf{1.5s} & 76m & \textbf{160.4k} & 
	901 & \textbf{0.6s} & 96m & \textbf{161.6k} &
	605 & \textbf{1.4s} & 159m & \textbf{165.7k} \\
Addresses & 155 & 5m & 5.4k & 
	181 & \textbf{3.9s} & 4m & \textbf{6.5k} & 
	182 & \textbf{3.9s} & 7m & \textbf{6.5k} &
	126 & \textbf{3.9s} & 12m & \textbf{7.3k} \\
JMLR & 580 & 8m & 29.2k & 
	583 & \textbf{5.4s} & 67m & \textbf{37.2k} & 
	593 & \textbf{6.5s} & 87m & \textbf{37.7k} & 
	334 & \textbf{5.6s} & 420m & \textbf{40.9k} \\
Moby & 231 & 46m & 9.6k &
	231 & \textbf{3m} & 23m & \textbf{10.9k} & 
	328 & \textbf{270s} & 39m & \textbf{10.9k}	 & 
	224 & \textbf{20.3s} & 66m & \textbf{12.5k}	\\ 
\bottomrule
\end{tabular}
\end{table*}

\begin{figure}[t]
\includegraphics{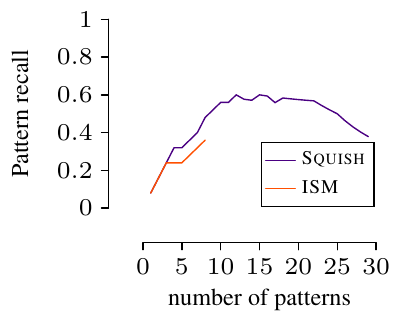}
\includegraphics{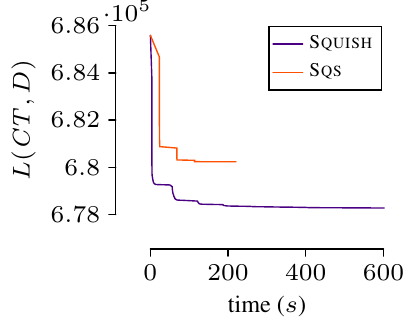}
\caption{(left) Recall of interleaving patterns (higher is better) for \SQUISH and \ISM on \textit{Parallel}. 
    (right) Runtime of \SQUISH and \SQS in seconds vs.\ encoded length of databases for \textit{Addresses} (lower is better).
    }\label{fig:extm}
    \label{fig:pr}  
\end{figure}

\begin{table}[t]
\centering
\caption{Sample choicisodes discovered by \SQUISH in the \emph{Addresses} and \emph{JMLR} datasets.}\label{tbl:cexamples}
\begin{tabular}{ll}
\toprule
& \textbf{Presidential Addresses} \\
 \midrule
1. & [coordin. $|$ execut.]  branch  govern \\

2. & fellow  [citizen $|$ american $|$ countrymen]  \\

3. & [discharg $|$ perform $|$ commenc]  duti \\

4. & god  [bless $|$ help] \\

5. & [exercise $|$ grant $|$ balanc]  power \\

6. & power  [grant $|$ vest] \\

7. & [eighteenth $|$ fifteenth $|$ fourteenth]  amendment \\

8. & [guard $|$ war]  against \\
\midrule
& \textbf{JMLR Abstracts} \\
\midrule
1. &[high $|$ curse $|$ low]  dimension. \\
 2. & [empirical $|$ structural]  risk  minimisation \\
 3. & [independent $|$ principle]  component  analysis \\
 4. & paper  [proposes $|$ presents]  new \\
 5. &  [Mahalanobis $|$ edit $|$ Euclidean $|$ pairwise]  distance  \\
 6. & [data $|$ train]  set  \\
 7. & [conditional $|$ Markov]  random  field   \\
 8. & [gradient $|$ coordinat.]  descent   \\
  \bottomrule
\end{tabular}
\end{table}

We further consider five real data sets. \textit{Gazelle} is click-stream data from an e-commerce website~\cite{kohavi:00:kddcup}. The \textit{Sign} database is a list of American sign language utterances~\cite{papapetrou2005discovering}. To allow for interpretability we also consider text data. Here the events are the (stemmed) words in the text, with stop words removed. \textit{Addresses} contains speeches of American presidents. \textit{JMLR} contains abstracts from the Journal of Machine Learning research, and \textit{Moby} is the famous novel Moby Dick by Herman Melville.

\subsubsection*{Synthethic Data}
As a sanity check we first compare to \SQS considering \emph{only} serial episodes and \emph{not} allowing interleaving or nesting. We find that in this setting \SQUISH performs on par with \SQS in terms of recovering non-interleaving patterns from synthetic data; like \SQS it correctly discovers no patterns from \textit{Indep}, it recovers all patterns from \textit{Plant-10}, and recovers 45 patterns exactly from \textit{Plant-50} and fragments of the remaining 5, but does so approximately ten times faster than \SQS.

To investigate how well \SQUISH retrieves interleaving patterns, we consider the \textit{Parallel} dataset, and compare to \ISM. (We also considered \SQS but found it did not finish within a day.) To make the comparison fair, we restrict ourselves again to serial episodes, but now do allow for interleaving and nesting. We measure success in terms of \emph{pattern recall}. That is, given a set of patterns $\mathcal{P}$ and a set of target patterns $\mathcal{T}$, we consider the set $\mathcal{T}$ as the data and cover it with $\mathcal{P}$ (not allowing for gaps). The pattern recall is the ratio of the total number covered events in $\mathcal{T}$  to the maximum of the total number of events in $\mathcal{T}$ or $\mathcal{P}$. 

We give the results in Fig.~\ref{fig:pr}. We find that \SQUISH obtains much higher recall scores than \ISM. Inspecting the results, we see that \SQUISH discovers large fragments of each pattern, whereas \ISM retrieves only eight small patterns, most of length 2, and hence does not reconstruct the generating set of patterns well.

\subsubsection*{Real data}

Next we evaluate \SQUISH on real data. We compare to \SQS in terms of number of patterns, achieved compression, and runtime. 
We consider three different configurations, 1) disjoint covers of only serial episodes, 2) allowing interleaving and nesting of serial episodes, and 3) allowing interleaving and nesting of serial episodes and choicisodes.
We give the results in Table~\ref{tbl:dcresults}.

First of all, the $\SQS$-$t$ columns show that in all setups \SQUISH needs only a fraction of the time---up to three orders of magnitude less---to discover a model that is at least good as what \SQS returns. To fully converge, \SQUISH and \SQS take roughly the same amount of time for the disjoint setting, as well as when we do allow interleaving. 
However, when converged \SQUISH discovers models with much better compression rates, i.e.\ with much higher $\Delta L$, than \SQS does. $\SQUISH$ is also significantly faster than $\ISM$, taking only 87 minutes instead of 259 on the \textit{JMLR} database, and on \textit{Gazelle} \SQUISH requires only 96 instead of 680 minutes.

\SQUISH performs best when we consider our richest description language, allowing both interleaving and choicisodes, discovering much more succinct models that obtain much better scores than if we restrict ourselves. For example, for \textit{Gazelle}, with choicisodes enabled \SQUISH needs only 605 instead of 901 patterns to achieve a $\Delta L$ of $165.7$k instead of $161.6$k. Overall, we observe that many choicisodes form semantically coherent groups. We present a number of exemplar choisisode patterns in Table \ref{tbl:cexamples}. Interesting examples include: data-set and training-set from \textit{JMLR}, god-bless and god-help from Address, cape-horn and cape-cod from \textit{Moby}. 

Last, but not least, we report on the convergence of $L(\CT,D)$, the encoded length of the database, over time for both \SQUISH and \SQS in Fig.~\ref{fig:extm}. Both algorithms estimate batches of candidates, and test them one by one tests. We see that the initial candidates are highly effective on increasing compression gain. Candidates generated in the latter iterations lead to only little increase in compression gain. This leads to the possibility of executing $\SQUISH$ based upon a time budget, as an any-time algorithm.

\section{Conclusion}\label{sec:concl}
We considered summarising event sequences. Specifically, we aimed at discovering sets of patterns that capture rich structure in the data. We considered interleaved, nested, and partial pattern occurrences. We proposed the algorithm $\FINDWIN$ to efficiently search for pattern occurrences and the greedy algorithm $\GREEDYCOVER$  for efficiently covering the data. Experiments show that \SQUISH works well in practice, outperforming the state of the art by a wide margin in terms of scores and speed, while discovering pattern sets that are both more succinct and easier to interpret.

As future work we are considering parallel episodes, patterns where certain events are un-ordered e.g.\ $a \left\{ b,c \right\} d$~\cite{mannila:97:discovery}. 
Discovering such structure presents a significant computational challenges and requires novel scores and algorithms.

\section*{Acknowledgements}
Apratim Bhattacharyya and Jilles Vreeken are supported by the Cluster of Excellence ``Multimodal Computing and Interaction'' within the Excellence Initiative of the German Federal Government.

\providecommand{\noopsort}[1]{}

\appendix

\section{Appendix}\label{sec:apx}
\subsection{Estimating Candidates}\label{sec:apx:est}
Here we describe our heuristic strategy for finding new candidates of the form $XY$ as in Sec.~\ref{subsec:gencan}.
First, we need two crucial observations.

\vspace{0.5em}
\noindent{\textbf{Constant Time Difference Estimation}} Given a database $D$ and an cover $C$. Let $P$ and $Q$ be two patterns. Let $V = \left\{v_{1},...,v_{N}\right\}$ and $W = \left\{ w_{1},...,w_{N}\right\}$ be two set of windows for $P$ and $Q$, respectively. Both $V$ and $W$ occur in $C$. Each of these windows $v_{i}$ and $w_{i}$ occur in the same sequence. Given the start positions and end positions of the pattern in sequence $k_{i}$,  we can write them as $v_i = (a_i, b_i, P, k_i)$ and $w_i = (c_i, d_i, Q, k_i)$. Let $U$ be the set of windows produced by combining them, $U = {(a_{1},d_{1},R,k_1),...,(a_N,d_N,R,k_N)}$. Let the windows in $U$ be disjoint and the windows in $U$ be disjoint with the windows in $C \setminus (V \cup W )$. Then the difference $L(D,C \cup U \setminus (V \cup W)) - L(D,C)$ depends only $N$, $gaps(V)$, $gaps(W)$, and $gaps(U)$ and can be computed in constant time from these values.
\vspace{0.5em}

\vspace{0.5em}
\noindent{\textbf{Shorter Windows in Optimal Cover}} Given a database $D$ and an cover $C$. Let $v = (i, j, X, k) \in C$. Assume that there exists a window $S\left[a, b\right]$ containing $X$ such that $w = (a, b, X, l)$ does not overlap with any window in $C$ and $b - a < j - i$. Then $C$ is not an optimal cover.
\vspace{0.5em}

We refer the reader to~\cite{tatti:12:sqs} for detailed proofs.

\begin{algorithm2e}[t]
\small
\Input{database $D$, cover $C$, and pattern $X$}
\Output{pattern $\mathit{XY}$ with low $L(D, \CT \cup \mathit{XY})$}

\ForEach{$Y \in \CT$} {
	$V_Y \leftarrow \emptyset$;
	$W_Y \leftarrow \emptyset$;
	$U_Y \leftarrow \emptyset$;
	$d_Y \leftarrow 0$\;
}

$T \leftarrow \emptyset$\;
\ForEach {window $v$ of $X$ in cover $C$} {
	$(a, b, X, k) \leftarrow v$\;
	$d \leftarrow $ end index of window following $v$ in $C$\;
	$\mathit{t} \leftarrow (v, d, 0)$; $l(t) \leftarrow d - a$\;
	add $t$ into $T$\;
}

\While {$T$ is not empty} {

	$t \leftarrow \arg\min_{u \in T} l(u)$\;

	$(v, d, s) \leftarrow t$; $a \leftarrow $ first index of $v$\;

	$w = (c, d, Y, k) \leftarrow$ active $w$ of $Y$ ending at $d$\;
	\If {$Y = X$ \AND (event at $a$ or $d$ is marked)} {
		delete $t$ from $T$\;
		\Continue\;
	}

	\If {$S_k[a, d]$ is a minimal window of $\mathit{XY}$} {
		add $v$ into $V_Y$\;
		add $w$ into $W_Y$\;
		add $(a, d, \mathit{XY}, k)$ into $U_Y$\;

		$d_Y \leftarrow \min(\diff(V, W, U; A) + s, d_Y)$\;
		\If {$|Y| > 1$} { $s \leftarrow s + \gain(w)$\;}
		\If {$Y = X$} {
			mark the events at $a$ and $d$\;
			delete $t$ from $T$\;
			\Continue\;
		}
	}
	\eIf {$w$ is the last window in the sequence} {
		delete $t$ from $T$\;
	}{
		$d \leftarrow$ end index of the active $w'$ following $w$\;
		update $t$ to $(v, d, s)$ and $l(t)$ to $d - a$\;
	}
}
\caption{$\ESTIMATE(X,C,D)$. Heuristic for finding pattern $\mathit{XY}$ with low $L(D,\CT \cup \mathit{XY})$}\label{estimate}
\end{algorithm2e}

We present our heuristic procedure $\ESTIMATE$ as pseudo-code in Algorithm \ref{estimate}. In this algorithm, given pattern $X$ and a cover $C$, for a possible extension $Y$, we enumerate the windows of $XY$ from the shortest to the longest. These windows are constructed by combining two windows in the cover $C$. We maintain the sets $V_Y$, $W_Y$ and $U_Y$ (line 1), containing windows of $X$, windows of $Y$ (to be combined together), and new windows of $XY$ (resulting from the combination) respectively. We do this for every possible extension $Y$ in the code table. At each step we compute the difference in code length of using these windows instead. We maintain $d_Y$ to store this difference. By the observation \textbf{Constant Time Difference Estimation}, this can be done in constant time. We prefer patterns $XY$ which are frequently occurring, with more fills than other meta stream characters. Thus, we want to find shorter windows of $XY$ first. Such a set of windows $U$ could potentially lead to a estimated decrease in code length. Therefore, to ensure that we find shorter windows first and efficiency, we search for all windows (all possible $Y$) simultaneously using a priority queue $T$ and look only at windows in the cover $C$. For each window of $X$ in the cover $C$, we look at windows after it to construct windows $XY$ ($Y$ is the pattern of the window following the window of $X$). We initialize the priority queue $T$ with these windows (line 4-9), sorted based on length. At each step of the candidate generation algorithm, we retrieve once such window of $XY$ from the priority queue $T$ (line 11) add it to our list $U_{Y}$ of windows of $XY$ and estimate the change in code length (line 22). As we do not allow overlaps, we need to ensure that windows in $U_{Y}$ are not overlapping. If a window of $XY$ overlaps with any other window in $C$, we cannot use both of these windows at the same time. We take this into account by subtracting the $gain(w)$ of this window $w$ overlapping with the window of $XY$ (line 23)~\cite{tatti:12:sqs}. The $gain(w)$ of a window $w$ if a upper bound on the bits gained by encoding the events in the database with this window vs.\ encoding them as singletons. We define the gain as in \cite{tatti:12:sqs} for a window $w$ of the pattern $Y$ ($S_{k}[i,j]$),

\begin{align*}
	\gain(w) = & -L(\code_{p}(X)) - (j - i - \abs{X})L(\code_{g}(X)) \\
	&  - (\abs{X} - 1)L(\code_{f}(X)) + \sum_{x \in X} L(\code_{p}(x))\quad. \\
\end{align*}

Overlapping could also happen if $Y = X$. So we simply check if the adjacent scans have already used these two instances of $X$ for creating a window for pattern $XX$ (line 25). We now extend our search by looking at the window following the currently considered window of $Y$ in the cover $C$ (line 34). As we allow interleaving and nesting in our covers, we also look at possible windows $Y$ occurring inside or interleaved with windows of other patterns. That is, we look at singletons inside gaps of windows. For each window $X$ in the cover $C$, we look at all windows following it, until we reach the window of $X$ or the end of the cover.

\subsection{Pruning the Code Table}\label{sec:apx:prune}
Here, we present the algorithm we use to prune the code table $\CT$, used at line 12 of $\SQUISH$ as pseudo-code in Algorithm \ref{prune}.

\begin{algorithm2e}[ht]
\Input{pattern set $\mathcal{P}$, database $D$}
\Output{pruned pattern set $\mathcal{P}$\;}

\ForEach{$X \in \mathcal{P}$} {
	$\CT \leftarrow$ code table corresponding to $\GREEDYCOVER(D, \mathcal{P})$\;
	$\CT' \leftarrow$ code table obtained from $\CT$ by deleting $X$\;
	$g \leftarrow \sum_{w = (i, j, X, k) \in C} \gain(w)$\;

	\If {$g < L(\CT) - L(\CT')$} {
		\If {$L(D, \mathcal{P} \setminus X) < L(D, \mathcal{P})$} {
			$\mathcal{P} \leftarrow \mathcal{P} \setminus X$\;
		}
	}
}
\caption{$\PRUNE(\mathcal{P},D)$}\label{prune}
\end{algorithm2e}


\begin{thebibliography}{10}

\bibitem{achar2012discovering}
A.~Achar, S.~Laxman, R.~Viswanathan, and P.~Sastry.
\newblock Discovering injective episodes with general partial orders.
\newblock {\em Data Min.\ Knowl.\ Disc.}, 25(1):67--108, 2012.

\bibitem{agrawal:94:fast}
R.~Agrawal and R.~Srikant.
\newblock Fast algorithms for mining association rules.
\newblock In {\em VLDB}, pages 487--499, 1994.

\bibitem{bertens:16:ditto}
R.~Bertens, J.~Vreeken, and A.~Siebes.
\newblock Keeping it short and simple: Summarising complex event sequences with
  multivariate patterns.
\newblock In {\em KDD}, pages 735--744, 2016.

\bibitem{fowkes:16:ism}
J.~Fowkes and C.~Sutton.
\newblock A subsequence interleaving model for sequential pattern mining.
\newblock In {\em KDD}, 2016.

\bibitem{grunwald:07:book}
P.~Gr\"{u}nwald.
\newblock {\em The Minimum Description Length Principle}.
\newblock MIT Press, 2007.

\bibitem{kohavi:00:kddcup}
R.~Kohavi, C.~Brodley, B.~Frasca, L.~Mason, and Z.~Zheng.
\newblock {KDD-Cup} 2000 organizers' report: Peeling the onion.
\newblock {\em SIGKDD Explor.}, 2(2):86--98, 2000.
\newblock http://www.ecn.purdue.edu/KDDCUP.

\bibitem{lam:12:gokrimp}
H.~T. Lam, F.~M\"{o}rchen, D.~Fradkin, and T.~Calders.
\newblock Mining compressing sequential patterns.
\newblock In {\em SDM}, 2012.

\bibitem{laxman2007fast}
S.~Laxman, P.~Sastry, and K.~Unnikrishnan.
\newblock A fast algorithm for finding frequent episodes in event streams.
\newblock In {\em KDD}, pages 410--419. ACM, 2007.

\bibitem{vitanyi:93:book}
M.~Li and P.~Vit\'{a}nyi.
\newblock {\em An Introduction to Kolmogorov Complexity and its Applications}.
\newblock Springer, 1993.

\bibitem{mannila:97:discovery}
H.~Mannila, H.~Toivonen, and A.~I. Verkamo.
\newblock Discovery of frequent episodes in event sequences.
\newblock {\em Data Min.\ Knowl.\ Disc.}, 1(3):259--289, 1997.

\bibitem{papapetrou2005discovering}
P.~Papapetrou, G.~Kollios, S.~Sclaroff, and D.~Gunopulos.
\newblock Discovering frequent arrangements of temporal intervals.
\newblock In {\em ICDM}, pages 354--361. IEEE, 2005.

\bibitem{pei2006discovering}
J.~Pei, H.~Wang, J.~Liu, K.~Wang, J.~Wang, and P.~S. Yu.
\newblock Discovering frequent closed partial orders from strings.
\newblock {\em IEEE TKDE}, 18(11):1467--1481, 2006.

\bibitem{petitjean:16:skopus}
F.~Petitjean, T.~Li, N.~Tatti, and G.~I. Webb.
\newblock Skopus: Mining top-k sequential patterns under leverage.
\newblock {\em Data Min.\ Knowl.\ Disc.}, 30(5):1086--1111, 2016.

\bibitem{rissanen:78:mdl}
J.~Rissanen.
\newblock Modeling by shortest data description.
\newblock {\em Automatica}, 14(1):465--471, 1978.

\bibitem{rissanen:83:integers}
J.~Rissanen.
\newblock A universal prior for integers and estimation by minimum description
  length.
\newblock {\em Annals Stat.}, 11(2):416--431, 1983.

\bibitem{smets:12:slim}
K.~Smets and J.~Vreeken.
\newblock \textsc{Slim}: Directly mining descriptive patterns.
\newblock In {\em SDM}, pages 236--247. SIAM, 2012.

\bibitem{tatti:15:epirank}
N.~Tatti.
\newblock Ranking episodes using a partition model.
\newblock {\em Data Min.\ Knowl.\ Disc.}, 29(5):1312--1342, 2015.

\bibitem{tatti:11:multievent}
N.~Tatti and B.~Cule.
\newblock Mining closed episodes with simultaneous events.
\newblock In {\em KDD}, pages 1172--1180, 2011.

\bibitem{tatti:12:clsepi}
N.~Tatti and B.~Cule.
\newblock Mining closed strict episodes.
\newblock {\em Data Min.\ Knowl.\ Disc.}, 25(1):34--66, 2012.

\bibitem{tatti:12:sqs}
N.~Tatti and J.~Vreeken.
\newblock The long and the short of it: Summarizing event sequences with serial
  episodes.
\newblock In {\em KDD}, pages 462--470. ACM, 2012.

\bibitem{vereshchagin:03:kolmo}
N.~Vereshchagin and P.~Vitanyi.
\newblock Kolmogorov's structure functions and model selection.
\newblock {\em IEEE TIT}, 50(12):3265-- 3290, 2004.

\bibitem{vreeken:11:krimp}
J.~Vreeken, M.~{van Leeuwen}, and A.~Siebes.
\newblock \textsc{Krimp}: Mining itemsets that compress.
\newblock {\em Data Min.\ Knowl.\ Disc.}, 23(1):169--214, 2011.

\bibitem{wang:04:bide}
J.~Wang and J.~Han.
\newblock Bide: Efficient mining of frequent closed sequences.
\newblock {\em ICDE}, 0:79, 2004.

\end{thebibliography}
\end{document}